\documentclass{article}
\usepackage[utf8]{inputenc}
\usepackage{color}
\usepackage{threeparttable}
\usepackage{graphicx}
\usepackage{authblk}
\usepackage{hyperref}
\usepackage{graphicx}
\usepackage{booktabs}
\usepackage{subcaption}
\usepackage{multirow}
\usepackage[table,xcdraw]{xcolor}
\usepackage{hyperref}

\title{\bfseries{HarDNet-MSEG: A Simple Encoder-Decoder Polyp Segmentation Neural Network that Achieves over 0.9 Mean Dice and 86 FPS}}
\author[ ]{Chien-Hsiang Huang}
\author[ ]{Hung-Yu Wu}
\author[ ]{Youn-Long Lin}
\affil[ ]{Department of Computer Science, National Tsing Hua University}
\affil[ ]{\small{\{james128333, a9778875\}@gmail.com}, \ \ ylin@cs.nthu.edu.tw}
\date{}

\begin{document}
\maketitle

\begin{center}
    \begin{figure}[htbp]
    \centering
    \includegraphics[scale=0.3]{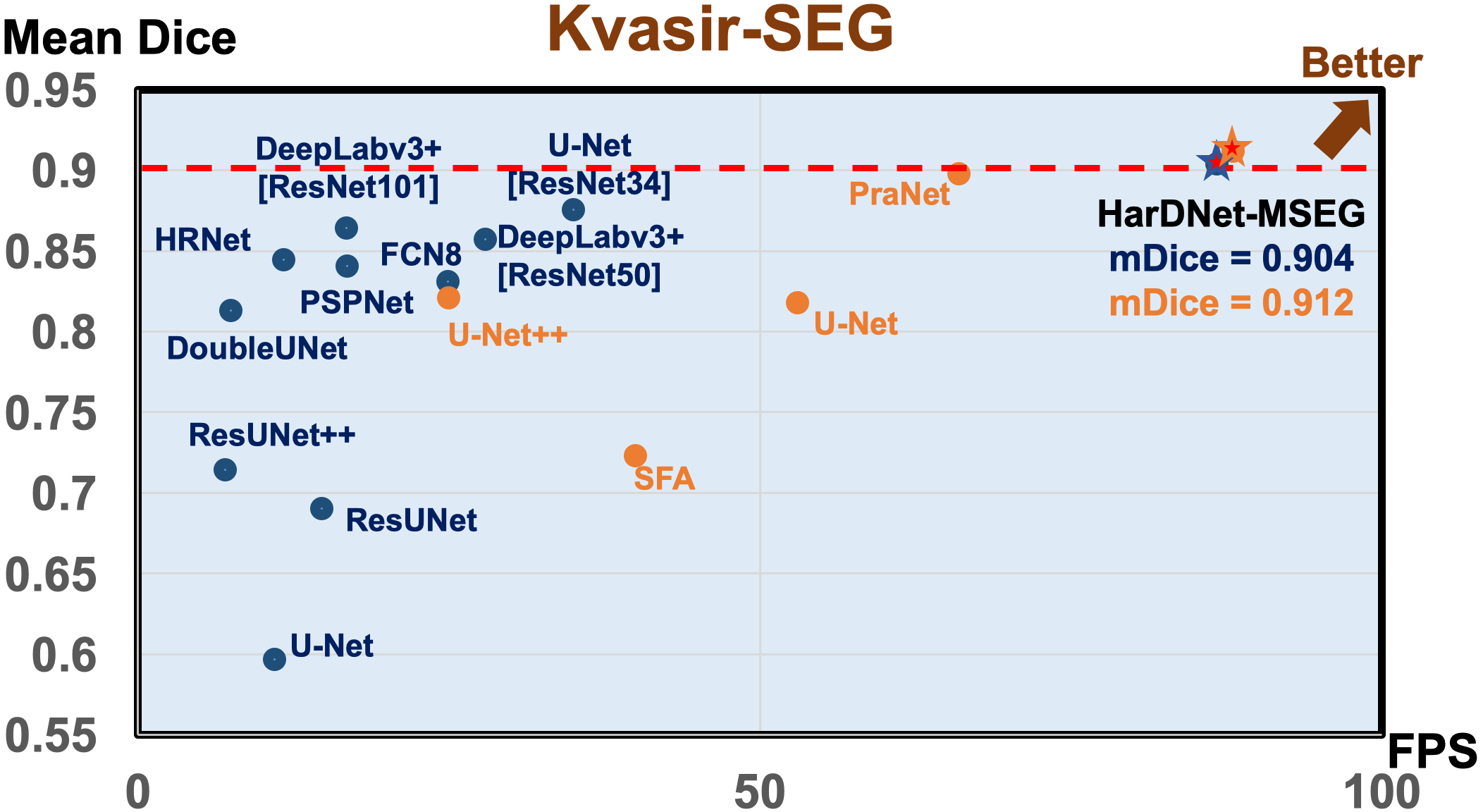}
    \caption{Mean Dice accuracy vs frame rate running on a GeForce RTX 2080 Ti GPU as reported in \textcolor{blue}{\cite{jha2020real}(blue)} and \textcolor{orange}{\cite{pranet}(orange)}. HarDNet-MSEG is faster and more accurate than the SOTA (U-Net[ResNet34] and PraNet).}
    \label{fig:landscapes}
    \end{figure}
\end{center}

\begin{abstract}
    We propose a new convolution neural network called HarDNet-MSEG for polyp segmentation. 
    It achieves the SOTA in both accuracy and inference speed on five popular  datasets (Kvasir-SEG\cite{jha2020kvasir}, CVC-ColonDB\cite{auto2},  EndoScene\cite{vazquez2017benchmark}, ETIS-Larib Polyp DB\cite{silva2014toward} and CVC-Clinic DB\cite{bernal2015wm}). 
    For Kvasir-SEG, HarDNet-MSEG delivers \textbf{0.904 mean Dice} running at \textbf{86.7 FPS} on a GeForce RTX 2080 Ti GPU (showing in Figure 1).
    It consists of a backbone and a decoder. 
    The backbone is a low memory traffic CNN called HarDNet68\cite{chao2019hardnet}, which has been successfully applied to various CV tasks including image classification, object detection, multi-object tracking ,and semantic segmentation, etc. 
    The decoder part is inspired by the Cascaded Partial Decoder\cite{wu2019cascaded}, known for fast and accurate salient object detection. We have evaluated HarDNet-MSEG using those five popular datasets. The code and all experiment details are available at Github. {{\color[HTML]{FE0000} \footnotesize{\textbf{https://github.com/james128333/HarDNet-MSEG}}}}
\end{abstract}
    
\section{Introduction}
    \hspace*{0.5cm}
    The incidence of colorectal cancer (CRC) has been ranked third in the world for many years. 
    Therefore, how to prevent CRC is an important global issue. Studies have pointed out that 95\% of CRC is due to a colorectal adenomatous polyp. 
    The resection of colorectal adenomatous polyps can greatly reduce the incidence of CRC. 
    Therefore, it is very important to have a colonoscopy on a regular basis as well as early invention and treatment.
    
\vspace{4mm}
    At present, the best way to prevent CRC is by taking regular colonoscopy and undergo a polyp removal resection. 
    With the emergence and popularization of painless colonoscopy, people's acceptance of the examination is getting higher.
    However, the detection of polyps was performed manually by endoscopists in the past, which is a consuming task for human beings and greatly depends on the doctor’s experience and ability. 
    Early segmentation methods\cite{auto1,auto2,auto3} are based on extracting features such as color, patterns, etc., and then using a classifier to distinguish polyps from their surroundings. 
    However, this method still has a high rate of missed detection. The position, size, color, etc. of each polyp are different, so it is very difficult to segment them automatically and accurately.

\vspace{4mm}
    In recent years, CNN has grown rapidly with breakthrough growth in the application of various imaging tasks. The segmentation of polyp have also benefited\cite{fully1,fully2}. 
    For this task, FCN\cite{FCN,fully1,fully2}, U-Net\cite{y-net,u-net}, U-Net++\cite{unet1,unet2}, DoubleU-Net\cite{jha2020doubleu} and ResUNet\cite{jha2019resunet++, yang2019road, jha2020real} series, etc., have good results compared to the early methods. 
    Most polyp blocks can be segmented out well, but there are still many problems, such as the cutting of boundary areas and the lack of smaller blocks, as well as broken images in large areas. 
    Moreover, the inference time of these networks is usually long, and the training time is relatively time-consuming.

\vspace{4mm}
    We propose HarDNet-MSEG based on the backbone of HarDNet68\cite{chao2019hardnet}. 
    With a simple encoder-decoder\cite{badrinarayanan2017segnet} architecture, it achieves excellent accuracy and efficient inference time for related benchmarks such as Kvasir-SEG, CVC-ColonDB, etc.

\section{Related work}
    \hspace*{0.5cm}Since the emergence of LeNet\cite{lecun2015lenet} in 1998, CNN has grown rapidly and has been used in different computer vision fields. Among them, the task of image segmentation is widely used in medical imaging.
    
\vspace{4mm}
    In 2015, Long et al. first introduced fully convolutional networks (FCN)\cite{FCN} for the task of image segmentation. 
    An end-to-end trained convolutional neural network is used to classify each pixel in an image.
    Since then, the convolutional neural network has flourished in the field of image segmentation. 
    In the same year, U-Net\cite{u-net} introduced at MICCAI has been widely used in the field of medical imaging. 
    Through a fairly symmetrical U-shaped encoder-decoder architecture, combined with skip connections at different scales to integrate deep and shallow features, it has now become a baseline network architecture for most medical imaging semantic segmentation. 
    Then, the emergence of U-Net++\cite{unet1,unet2} expands the original U-shaped architecture. 
    With more skip connections and convolutions to achieve the effect of deep layer aggregation\cite{yu2018deep}. It solves the problem that edge information and small objects are easily lost due to deep network down-sampling and up-sampling.
  
    \vspace{4mm}
    In recent years, the use of a better CNN backbone, or the introduction of additional modules like spatial pyramid pooling\cite{he2015spatial}, attention modules\cite{chen2016attention,fu2019dual}, etc., have achieved very good results in medical imaging semantic segmentation. 
    Examples of the former include ResUNet\cite{yang2019road,jha2020real}, ResUNet++\cite{jha2019resunet++}, and DoubleU-Net\cite{jha2020doubleu}. 
    By integrating a better CNN backbone with a U-shaped structure, the entire network has a stronger recognition capability, a larger receiving domain, and multi-scale information integration. 
    The second is to insert additional modules, such as DoubleU-Net\cite{jha2020doubleu} uses ASPP\cite{chen2018encoder} between the encoder and the decoder, which helps to deal with different object scales and improve accuracy; PraNet\cite{pranet} adds an RFB\cite{liu2018receptive} module to skip connection to capture more visual information for features of different scales. 
    In recent years, attention has also been widely used in the field of computer vision, especially for semantic segmentation which requires detailed edge information at the pixel level. Examples include PraNet\cite{pranet}, PolypSeg\cite{zhong2020polypseg} and ABC-Net\cite{fang2020abc}. 
    After adding different context modules, they all get good results in medical imaging segmentation. 
    Context modules such as Spatial Attention Module\cite{chen2017sca} and Channel Attention Module\cite{chen2017sca} will reduce the inference speed, but on the other hand, they are very efficient in improving accuracy and making edge cutting more precise.

    \vspace{4mm}
    The HarDNet-MSEG we proposed uses HarDNet\cite{chao2019hardnet} as the backbone and is designed with an encoder-decoder architecture. 
    It has achieved the high accuracy of the current state of the art in CVC-ColonDB,  EndoScene, ETIS-Larib Polyp DB, CVCClinic DB, and Kvasir-SEG, and at the same time has an efficient inference speed. 
    In addition, we have also tried to add additional modules such as RFB, ASPP, Attention, etc. to our network architecture to further improve the accuracy.
\newpage


\begin{figure}[htb]
\centering
\includegraphics[scale=0.35]{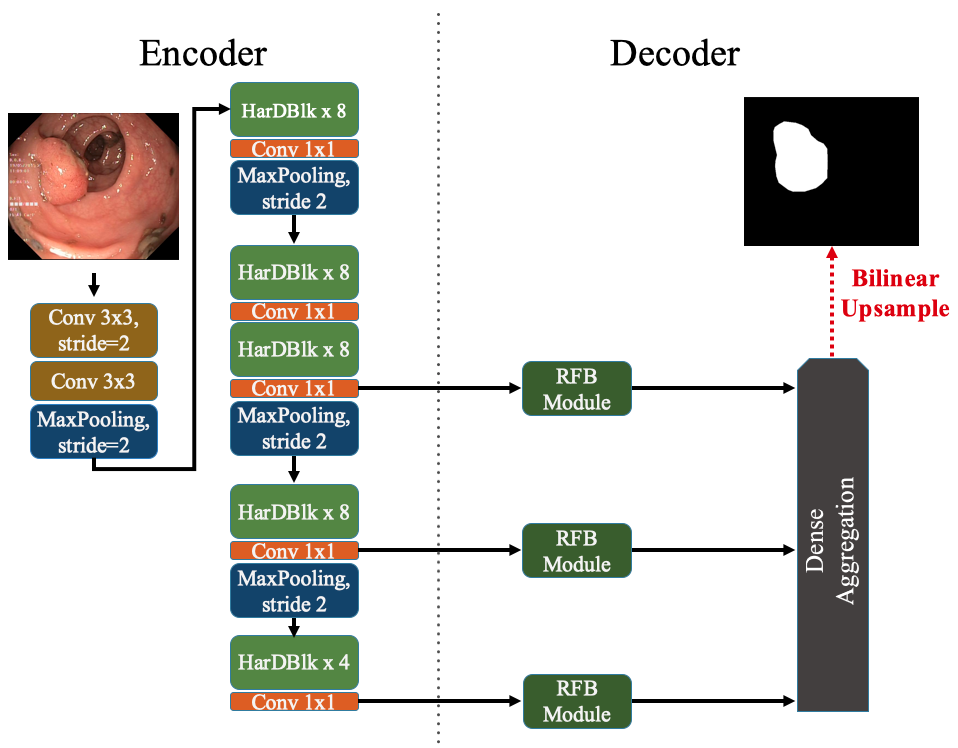}
\caption{
HarDNet-MSEG overview.
The encoder part consists of HarDNet68, 
and the decoder part is using partial decoder.}
\label{fig:HarDNet - MED}
\end{figure}

\begin{figure}[htb]
\centering
\includegraphics[scale=0.5]{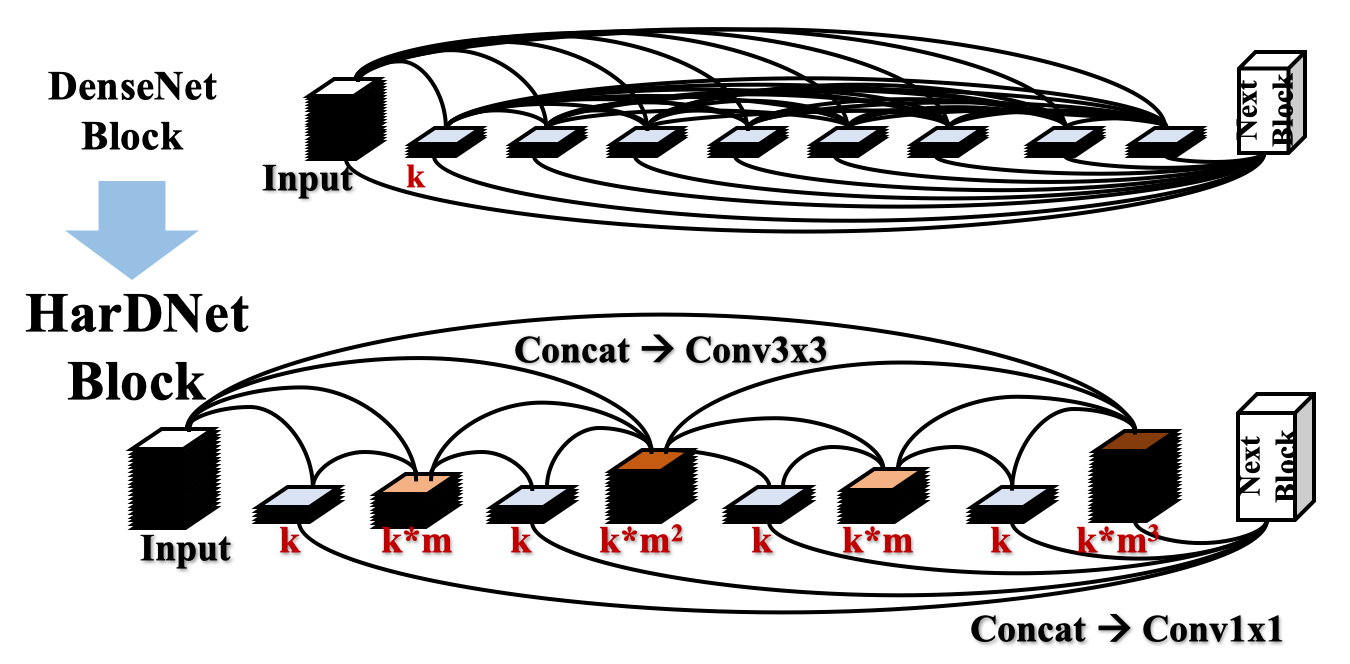}
\caption{HarDNet Block overview.}
\label{fig:hardblk}
\end{figure}

\section{HarDNet-MSEG}
    \vspace{4mm}
    \hspace*{0.5cm}Figure 2 depicts the architecture of our proposed HarDNet-MSEG. It consists of an encoder backbone and a decoder.
\subsection{Backbone : HarDNet}
    \vspace{4mm}
    \hspace*{0.5cm}HarDNet\cite{chao2019hardnet}, improved the original dense block of Densenet\cite{huang2017densely} are illustrated in Figure 3. Considering the impact of memory traffic on model design, it reduces shortcuts to increase the inference speed, and at the same time increases its channels’ width for the key layer to make up for the loss of accuracy. 
    It also uses a small amount of Conv1x1 to increase computational density. 

\vspace{4mm}
    Through this design, it not only achieves 30\% inference time reduction compared with DenseNet\cite{huang2017densely} and ResNet\cite{he2016deep}, also having higher accuracy on ImageNet\cite{deng2009imagenet}. 
    On the other hand, FC-HarDNet70\cite{huang2017densely} also reaches the state of the art in image segmentation on Cityscapes Dataset\cite{cordts2016cityscapes}. Therefore, we use HarDNet68 as the model backbone for Colorectal Polyps image semantic segmentation.

\subsection{Cascaded Partial Decoder}
    \vspace{4mm}
    \hspace*{0.5cm}Many well-known medical image segmentation networks are often modified based on the U-Net. 
    Our design also went in this direction at the beginning. 
    But based on the balance of the inference time and performance, we did not use HarDBlock (HarDBlk) in the Decoder part, which is different from FC-HarDNet.
    
    \vspace{4mm}
    We reference the Cascaded partial decoder \cite{wu2019cascaded}. 
    It found out that the shallow features have high resolution and occupy computing resources, and the deep information can also represent the spatial details of the shallow information relatively well. 
    So we decide to discard the shallow features and do more computing on the deeper layers’ features. 
    At the same time, the aggregation of feature maps at different scales can be achieved by adding appropriate convolution and skip connections.

\begin{figure}[htb]
\centering
\includegraphics[scale=0.53]{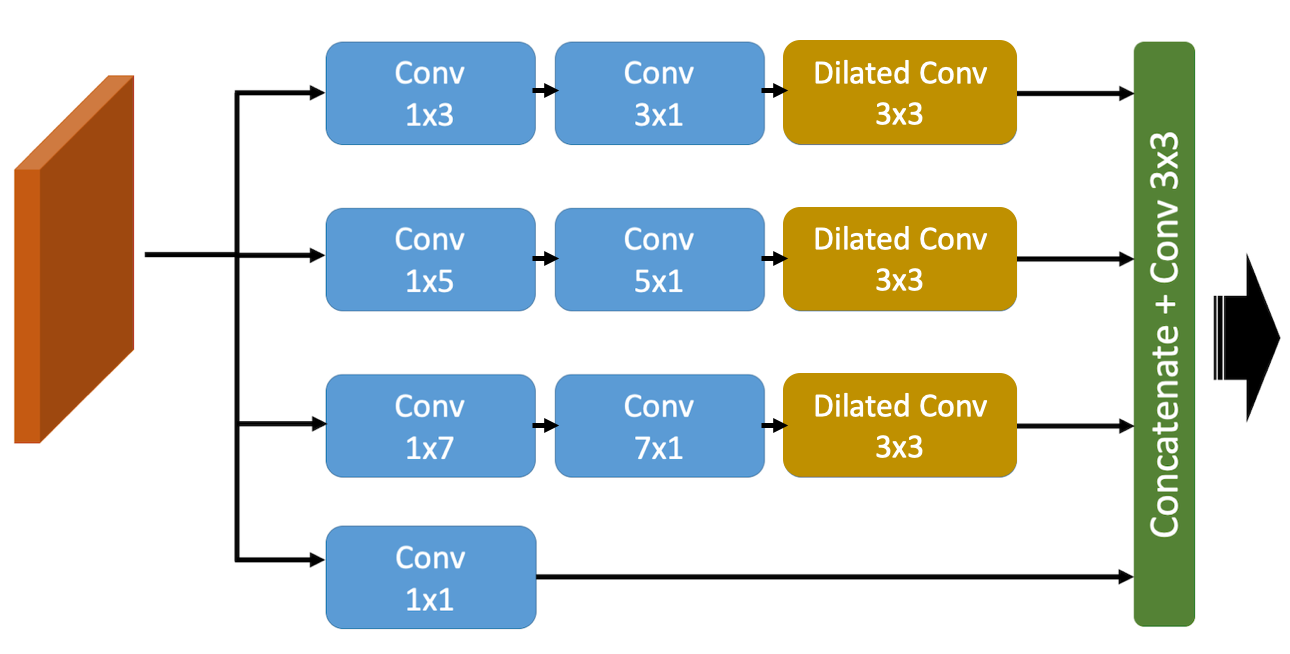}
\caption{RFB Module overview.}
\label{fig:rfb}
\end{figure}

\subsubsection{RFB Module} 
    \vspace{4mm}
    \hspace*{0.5cm}Figure 4 shows a Receptive Field Block\cite{liu2018receptive}. It can strengthen the deep features learned from a lightweight CNN backbone.
    By using multi-branch with different kernel size convolution and dilated convolution layers, it generates features with the different receptive fields. Afterwards, it applies a 1x1 convolution to merge these features and generate the final representation.
    
    \vspace{4mm}
    We add this module to the skip connection according to \cite{wu2019cascaded}, so that we could enlarge our receptive fields from each different resolutions’ feature maps.

\subsubsection{Dense Aggregation}

    \vspace{4mm}
    \hspace*{0.5cm}We perform aggregation by element-wise multiplication shown in Figure 5. 
    After up-sampling to the same scale, the feature is multiplied with another input feature of the corresponding scale. 

\begin{figure}[htb]
\centering
\includegraphics[scale=0.5]{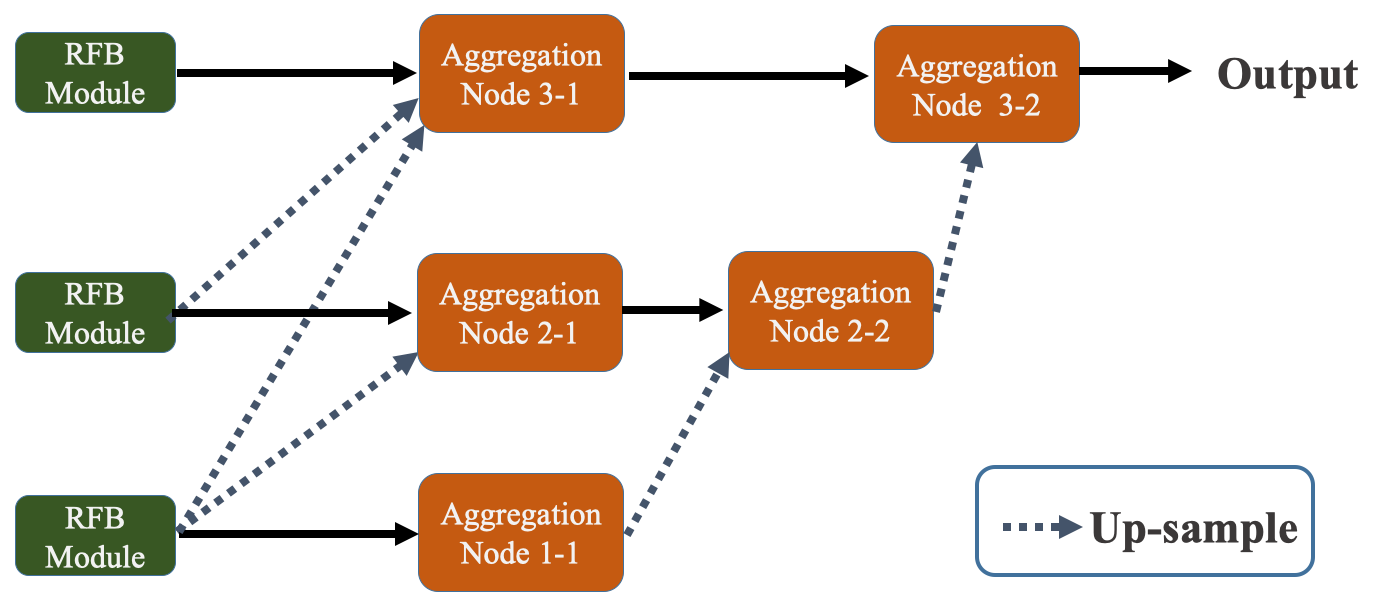}
\caption{Aggregation Module overview.}
\label{fig:AGG}
\end{figure}

\newpage

\section{Experiments}
    \hspace*{0.5cm}
    We used the training data from \cite{pranet} and \cite{jha2020real} for training because they have excellent performance in polyp segmentation. The training data and training methods used in the two articles are different. In order to reduce the variable factors, the training methods we use will refer to the methods proposed in the two articles respectively, and then compare the accuracy and inference speed with other models.

\begin{center}
\tabcolsep=3pt
\begin{table}[ht]
\resizebox{\textwidth}{!}{
\begin{tabular}{|c|c|c|c|c|c|c|c|c|}
\hline
              & \textbf{mIoU}   & \textbf{mDice} & \textbf{F2-score} & \textbf{Precision} & \textbf{Recall} & \textbf{Overall Acc.} & \textbf{FPS}   \\ \hline
U-Net                & 0.471 & 0.597    & 0.598    & 0.672    & 0.617 & 0.894       & 11    \\ \hline
ResUNet           & 0.572 & 0.690    & 0.699   & 0.745    & 0.725 & 0.917       & 15  \\ \hline
ResUNet++             & 0.613 & 0.714    & 0.720   & 0.784    & 0.742 & 0.917       & 7     \\ \hline
FCN8            & 0.737 & 0.831     & 0.825   & 0.882   & 0.835 & 0.952       & 25  \\ \hline
HRNet                  & 0.759 & 0.845    & 0.847   & 0.878    & 0.859 & 0.952       & 12  \\ \hline
DoubleUNet       & 0.733 & 0.813    & 0.820   & 0.861    & 0.840 & 0.949       & 7.5   \\ \hline
PSPNet        & 0.744 & 0.841    & 0.831   & 0.890    & 0.836 & 0.953       & 17  \\ \hline
DeepLabv3+[ResNet50]     & 0.776 & 0.857    & 0.855   & 0.891    & 0.8616 & 0.961       & 28  \\ \hline
DeepLabv3+[ResNet101]    & 0.786 & 0.864    & 0.857    & 0.906    & 0.859 & 0.961       & 17 \\ \hline
U-Net[ResNet34]          & 0.810   & 0.876    & 0.862   & {\color[HTML]{FE0000} \textbf{0.944}}    & 0.860 & 0.968       & 35    \\ \hline
{\color[HTML]{FE0000} \textbf{HarDNet-MSEG}}  & {\color[HTML]{FE0000} \textbf{0.848}} & {\color[HTML]{FE0000} \textbf{0.904}} & {\color[HTML]{FE0000} \textbf{0.915}} & 
0.907 & 
{\color[HTML]{FE0000} \textbf{0.923}} & 
{\color[HTML]{FE0000} \textbf{0.969}} & 
{\color[HTML]{FE0000} \textbf{86.7}} \\ \hline
\end{tabular}
}
\renewcommand{\arraystretch}{1.3}
        \caption{Quantitative results on Kvasir dataset (training/testing split:880/120). Showing the performance of different metrics and inference speed evaluating on GeForce RTX 2080 Ti GPU. Others evaluation scores are refer from \cite{jha2020real}.}
\end{table}
\end{center}

\subsection{Dataset }
    \vspace{4mm}
    \hspace*{0.5cm}We used the datasets proposed in the two papers mentioned earlier, namely Kvasir-SEG, CVC-ColonDB,  EndoScene, ETIS-Larib Polyp DB, and CVC-Clinic DB. And we will make a detailed comparison with other SOTA models on these datasets.
    
\subsection{Training setting and policy}
\vspace{4mm}
\hspace*{0.5cm}The two articles are based on different splitting method of training data, so we made two different experiments on each training setting to compare, and the details of the experiments will be explained below.

    \vspace{4mm}
    For \cite{jha2020real}, 880 images of Kvasir-SEG is used for training, and the other 120 images are used for testing. It does use augmentations like random rotation, horizontal flip, vertical flip. Our training input size is 512x512. We train our model with SGD optimizer for 100 epochs and the learning rate is set  to 1e-2. The results comparing to \cite{jha2020real} is in Table 1. HarDNet-MSEG shows the greatest accuracy on most metrics, and the inference speed is much faster than others. 

    \vspace{4mm}
    In \cite{pranet}, 1450 training images without any augmentation is used, including 900 images in Kvasir-SEG and 550 images in CVC-ClinicDB. And the testing set has 5 datasets mentioned above. Our training input size is 312x312, We train our model with Adam optimizer for 100 epochs and the learning rate is set to 1e-4. The quantitative results of each 5 datasets are shown in Table 2 (Kvasir-SEG) and Table 3 (ETIS, CVC-ClinicDB, CVC-ColonDB and EndoScene). We achieve the best performance in mean Dice and mIoU on each dataset, with the fastest inference speed (88 FPS).

\begin{table}[htb]
\label{tab:my-table}
\resizebox{\textwidth}{!}{
\begin{tabular}{|l|l|l|l|l|l|l|l|}
\hline
  &
  \textbf{mDice} &
  \textbf{mIoU} &
  \textbf{wfm} &
  \textbf{Sm} &
  \textbf{MAE} &
  \textbf{maxEm} &
  \textbf{FPS} \\ \hline
 
U-Net &
  0.818 &
  0.746 &
  0.794 &
  0.858 &
  0.055 &
  0.893 &
  53 \\ \hline

U-Net++ &
  0.821 &
  0.743 &
  0.808 &
  0.862 &
  0.048 &
  0.910 &
  25 \\ \hline
ResUNet-mod &
0.791 &
  n/a &
  n/a &
  n/a &
  n/a &
  n/a &
n/a \\ \hline
ResUNet++ &
0.813 &
0.793 &
  n/a &
  n/a &
  n/a &
  n/a &
n/a \\ \hline

SFA &
  0.723 &
  0.611 &
  0.67 &
  0.782 &
  0.075 &
  0.849 &
  40 \\ \hline

PraNet &
  0.898 &
  0.840 &
  0.885 &
  0.915 &
  0.030 &
  0.948 &
  66 \\ \hline

{\color[HTML]{FE0000} \textbf{HarDNet-MSEG}} &
  {\color[HTML]{FE0000} \textbf{0.912}} &
  {\color[HTML]{FE0000} \textbf{0.857}} &
  {\color[HTML]{FE0000} \textbf{0.903}} &
  {\color[HTML]{FE0000} \textbf{0.923}} &
  {\color[HTML]{FE0000} \textbf{0.025}} &
  {\color[HTML]{FE0000} \textbf{0.958}} &
  {\color[HTML]{FE0000} \textbf{88}} \\ \hline
\end{tabular}
}
\renewcommand{\arraystretch}{2}
        \caption{Quantitative results on Kvasir, comparing with the SOTA. Using the same training script with the release code of PraNet. The inference speed is testing under 312x312 resolution on GeForce RTX 2080 Ti GPU.}
\end{table}

\begin{table}[htb]
\label{tab:my-tabletab}
\resizebox{\textwidth}{!}{
\begin{tabular}{|c|c|c|c|c|c|c|c|c|}
\hline
 &
  \multicolumn{2}{c|}{\textbf{ClinicDB}} &
  \multicolumn{2}{c|}{\textbf{ColonDB}} &
  \multicolumn{2}{c|}{\textbf{ETIS}} &
  \multicolumn{2}{c|}{\textbf{CVC-T}} \\ \cline{2-9} 
\multirow{-2}{*}{} &
  \textbf{mDice} &
  \textbf{mIoU} &
  \textbf{mDice} &
  \textbf{mIoU} &
  \textbf{mDice} &
  \textbf{mIoU} &
  \textbf{mDice} &
  \textbf{mIoU} \\ \hline
U-Net &
  0.823 &
  0.755 &
  0.512 &
  0.444 &
  0.398 &
  0.335 &
  0.71 &
  0.627 \\ \hline
U-Net++ &
  0.794 &
  0.729 &
  0.483 &
  0.410 &
  0.401 &
  0.344 &
  0.707 &
  0.624 \\ \hline
ResUNet-mod &
  0.779 &
  n/a &
  n/a &
  n/a &
  n/a &
  n/a &
  n/a &
  n/a \\ \hline
ResUNet++ &
  0.796 &
  0.796 &
  n/a &
  n/a &
  n/a &
  n/a &
  n/a &
  n/a \\ \hline
SFA &
  0.700 &
  0.607 &
  0.469 &
  0.347 &
  0.297 &
  0.217 &
  0.467 &
  0.329 \\ \hline
PraNet &
  0.899 &
  0.849 &
  0.709 &
  0.640 &
  0.628 &
  0.567 &
  0.871 &
  0.797 \\ \hline
{\color[HTML]{FE0000} \textbf{HarDNet-MSEG}} &
  {\color[HTML]{FE0000} \textbf{0.932}} &
  {\color[HTML]{FE0000} \textbf{0.882}} &
  {\color[HTML]{FE0000} \textbf{0.731}} &
  {\color[HTML]{FE0000} \textbf{0.660}} &
  {\color[HTML]{FE0000} \textbf{0.677}} &
  {\color[HTML]{FE0000} \textbf{0.613}} &
  {\color[HTML]{FE0000} \textbf{0.887}} &
  {\color[HTML]{FE0000} \textbf{0.821}} \\ \hline
\end{tabular}
}
\renewcommand{\arraystretch}{2}
        \caption{More results on CVC-ClinicDB, CVC-ColonDB, ETIS, and CVC-T, comparing with the SOTA. Among them, CVC-T is the testing data for EndoScene.}
\end{table}

\subsection{Metrics}
\hspace*{1.5cm}Mean Dice $=\frac{2*tp}{2*tp+fp+fn}$
\hspace*{1.2cm}mIoU $=\frac{tp}{tp+fp+fn}$
\vspace{4mm}\newline
\hspace*{2.2cm}Recall $=\frac{tp}{tp+fn}$
\hspace*{1.5cm}Precision $=\frac{tp}{tp+fp}$
\vspace{4mm}\newline
\hspace*{2.8cm}F2 $=\frac{5p*r}{4p+r}$
\hspace*{2.3cm}Acc. $=\frac{tp+tn}{tp+tn+fp+fn}$

\vspace{4mm}
\hspace*{0.5cm}We will mainly use Kvasir's official website as the basis for comparison, namely mean Dice and Mean IoU, but we will still use other metrics mentioned in these two articles for comparison so that we can show our advantages more clearly.

\subsection{Training and Inference Environment setting:
}
\vspace{4mm}
\hspace*{0.5cm}In order to show our advantage in speed, we respectively compare with other famous models. The platforms we use for evaluation is written below.

\vspace{2mm}{\footnotesize{\textbf{Intel i9-9900k CPU, GeForce RTX 2080 Ti, Pytorch: 1.6 and CUDA: 10.2}}}

\section{Conculsion}
\hspace*{0.5cm}HarDNet-MSEG achieved the SOTA in all five challenging datasets. It is the only network that has achieved over 0.90 mean Dice (0.912 comparing with \cite{pranet} and 0.904 comparing with \cite{jha2020real}) on Kvasir-SEG. And it is 1.3 times faster than PraNet and more than 2 times faster than other models. We achieve this with a simple encoder-decoder architecture without any attention module used in \cite{pranet} and \cite{fang2020abc}. See Figure 6 for some inference results of Kvasir-SEG. It shows that our model outputs better boundary and the prediction is more accurate.

\vspace{4mm}
Again, it shows that HarDNet\cite{chao2019hardnet} is a great and efficient backbone in not only classification and detection, but also medical imaging segmentation. We hope this study can help pushing the frontier of medical imaging and contribute to the application of CNN in this field.
\begin{figure}[htb]
\centering
\includegraphics[scale=0.25]{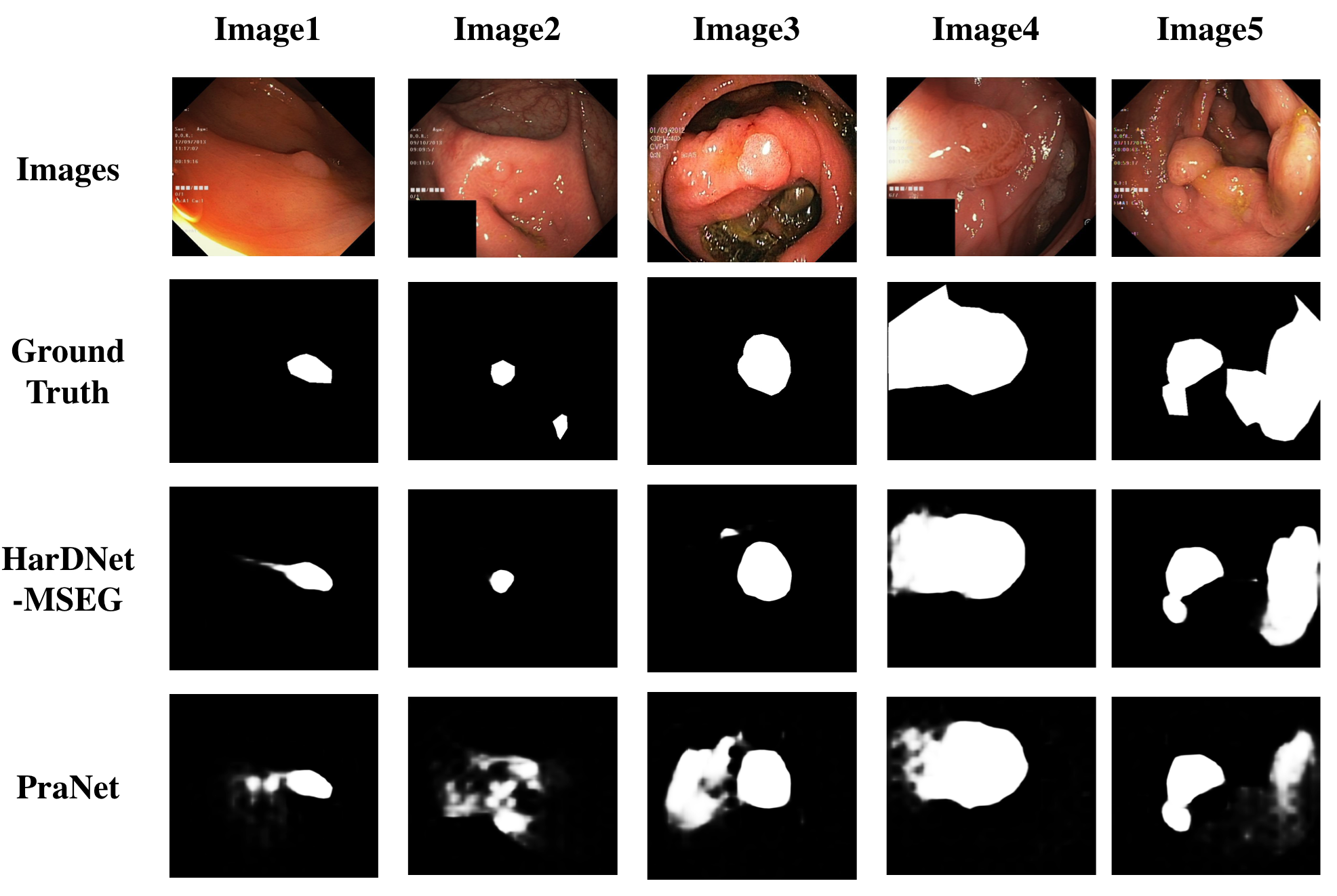}
\caption{Inference results of Kvasir-SEG.}
\label{fig:kva}
\end{figure}

\section*{Acknowledgements}
\hspace*{0.5cm}This research is supported in part by a grant from the
Ministry of Science and Technology (MOST) of Taiwan. We thank National Center for High-performance Computing (NCHC) for providing computational and storage resources. Without it this research is impossible. We would also like to thank Mr.Ping Chao for many fruitful discussions.


\bibliographystyle{plain}
\bibliography{ref}
\end{document}